\documentclass[10pt,twocolumn,letterpaper]{article}

\usepackage[pagenumbers]{cvpr} % To force page numbers, e.g. for an arXiv version

\usepackage[accsupp]{axessibility}  % Improves PDF readability for those with disabilities.
\usepackage{graphicx}
\usepackage{amsmath}
\usepackage{amssymb}
\usepackage{booktabs}
\usepackage{multirow}
\usepackage{algorithm}
\usepackage{algpseudocode}
\usepackage{color, colortbl}
\usepackage{subcaption}

\algnewcommand\Input{\item[\textbf{Input:}]}%
\algnewcommand\Output{\item[\textbf{Output:}]}%

\usepackage[pagebackref,breaklinks,colorlinks]{hyperref}

\definecolor{Gray}{gray}{0.9}

\usepackage[capitalize]{cleveref}
\crefname{section}{Sec.}{Secs.}
\Crefname{section}{Section}{Sections}
\Crefname{table}{Table}{Tables}
\crefname{table}{Tab.}{Tabs.}

\def\cvprPaperID{10478} % *** Enter the CVPR Paper ID here
\def\confName{CVPR}
\def\confYear{2023}

\begin{document}

\title{ViPLO: Vision Transformer based Pose-Conditioned Self-Loop Graph for Human-Object Interaction Detection}

\author{
Jeeseung Park$^1$ \qquad Jin-Woo Park$^{1,2}$ \qquad Jong-Seok Lee$^2$ \smallskip\\
{$^{1}$mAy-I Inc.,\,Seoul,\,Korea,\qquad $^{2}$Yonsei\,University, \,Korea} \\
{\tt\small \{jspark,\,jin\}@may-i.io \qquad jong-seok.lee@yonsei.ac.kr}
}

%\author{Jeeseung Park\\
%mAy-I Inc.\\
%Seoul, Korea\\
%{\tt\small jspark@may-i.io}
%\and
%Jin-Woo Park\\
%mAy-I Inc.\\
%Seoul, Korea\\
%{\tt\small 
%jin@may-i.io}
%\and
%Jong-Seok Lee\\
%Yonsei University\\
%Seoul, Korea\\
%{\tt\small 
%jong-seok.lee@yonsei.ac.kr}
%}
\maketitle

%%%%%%%%% ABSTRACT
\begin{abstract}
   Human-Object Interaction (HOI) detection, which localizes and infers relationships between human and objects, plays an important role in scene understanding. Although two-stage HOI detectors have advantages of high efficiency in training and inference, they suffer from lower performance than one-stage methods due to the old backbone networks and the lack of considerations for the HOI perception process of humans in the interaction classifiers. In this paper, we propose Vision Transformer based Pose-Conditioned Self-Loop Graph (ViPLO) to resolve these problems. First, we propose a novel feature extraction method suitable for the Vision Transformer backbone, called masking with overlapped area (MOA) module. The MOA module utilizes the overlapped area between each patch and the given region in the attention function, which addresses the quantization problem when using the Vision Transformer backbone. In addition, we design a graph with a pose-conditioned self-loop structure, which updates the human node encoding with local features of human joints. This allows the classifier to focus on specific human joints to effectively identify the type of interaction, which is motivated by the human perception process for HOI. As a result, ViPLO achieves the state-of-the-art results on two public benchmarks, especially obtaining a +2.07 mAP performance gain on the HICO-DET dataset. The source codes are available at \url{https://github.com/Jeeseung-Park/ViPLO}.

\end{abstract}

%%%%%%%%% BODY TEXT
\section{Introduction}
\label{sec:1}
Human-Object Interaction (HOI) detection aims to localize human and objects in the image, and classify interactions between them. A detected HOI instance is represented in the form of {\emph{$\langle human, object, interaction\rangle$}} triplet. HOI detection is steadily attracting attention in the computer vision field, due to the high potential for use in the downstream tasks such as action recognition \cite{yan2018spatial}, scene understanding \cite{xiao2018unified}, and image captioning \cite{you2016image}.

Existing HOI detectors can be divided into two categories: two-stage methods and one-stage methods. The former usually consists of three steps: 1) human and object detection with an off-the-shelf detector; 2) feature extraction for human and objects with ROI-Pooling; and 3) prediction of the interaction between human and objects with the extracted features \cite{chao:wacv2018, gao2018ican, gupta2019no,  wan2019pose, hou2021affordance, hou2020visual}.  One-stage HOI detection framework is first proposed in PPDM \cite{liao2020ppdm}, which directly predicts HOI triplets based on interaction points \cite{wang2020learning, liao2020ppdm} and union boxes \cite{kim2020uniondet}. Recent studies based on one-stage structure use the two-branch transformer, which predict HOI triplets with two sub-task decoders and matching processes \cite{liao2022gen, zhou2022human}. Nevertheless, these methods suffer from low training speed and large memory usage \cite{zhang2022efficient, zhu2020deformable}. 

%Recent one-stage methods follow DETR \cite{carion2020end}, which formulate HOI detection as a set prediction problem and predict triplets based on learnable queries \cite{kim2021hotr, tamura2021qpic, chen2021reformulating}.
%to get improved performance
%% HOI Transformers utilize single-branch methods at the beginning \cite{kim2021hotr, tamura2021qpic}, predicting HOI triplets with a single decoder. However, these kinds of one-branch methods suffer from the difference in feature presentation between object detection and interaction understanding, resulting in a sub-optimal solution. Recent studies show that two-branch methods which predict HOI triplets with two sub-task decoders and matching processes show better performance \cite{liao2022gen, zhou2022human}.  

\begin{figure*}[t]
\begin{center}

\includegraphics[width=0.75\linewidth]{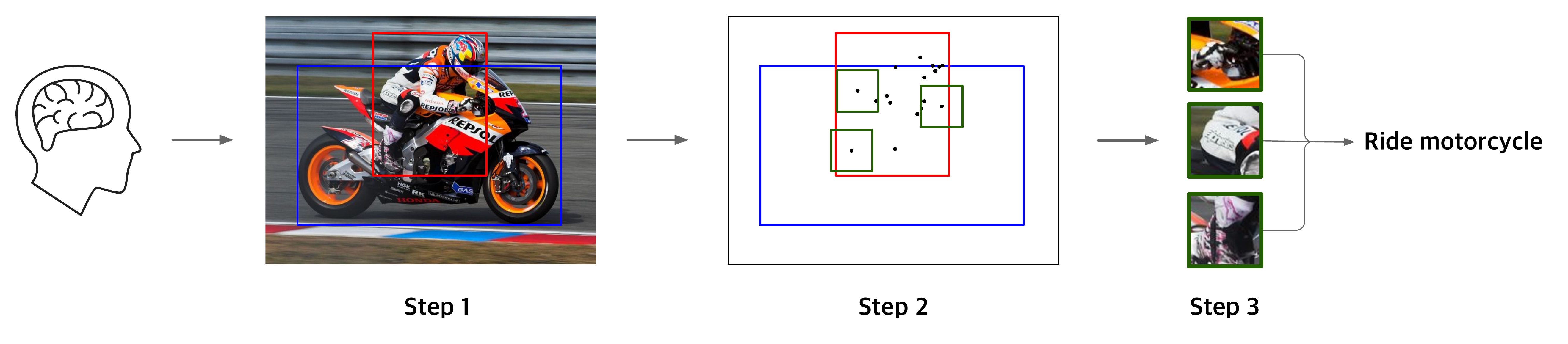}

\end{center}
\vspace{-5mm}
   \caption{The process of HOI recognition by humans. Humans first localize each person and object (Step 1), identify interactiveness using the spatial relationship and human pose (Step 2), and finally focus on specific human joints to recognize the type of interaction (Step 3). 
}
\vspace{-1mm}
\label{fig:human_perception}
\end{figure*} 

In contrast, two-stage HOI detectors show high training speed using pre-trained object detectors. They are also advantageous when the object bounding box is known in advance from the image. For example, when analyzing HOI events of fixed target objects, we can efficiently detect interaction
%using detected human region and fixed object region. 
using the fixed object regions without an extra object detection process. Moreover, interaction inference can be performed only for the desired human-object pair, which is also a big advantage compared to the one-stage methods in the application situation. Despite these advantages, two-stage methods have recently been less studied due to their low performance. In this paper, we focus on improving the performance of the two-stage approach, which can also enjoy the benefit of reduced computational and memory complexity. In particular, we improve the two important parts: feature extraction and interaction prediction.
% Thus, we consider improving the two-stage methods from the two aspects, feature extraction and interaction prediction. 
% 윗 문단의 마지막에서 두 번째 In this paper~

%Despite these advantages, two-stage methods suffer from low performance than one-stage methods.  
% 그래서 학계에서 외면받았다..?

%Despite these advantages, two-stage methods have recently been less studied due to their low performance. We consider improving two-stage methods by focusing on the second and third steps of the framework. 
Regarding the feature extraction part, we note that most of the two-stage methods rely on ResNet \cite{he2016deep} as backbone networks \cite{hou2020visual, hou2021affordance, zhang2021spatially}. Recently, the Vision Transformer (ViT) has emerged as a powerful alternative yielding state-of-the-art performance in various computer vision fields \cite{radford2021learning, touvron2021training, strudel2021segmenter}, and it also has a potential as an improved feature extractor for two-stage HOI detection. However, it is not straightforward to apply a conventional feature extraction method (e.g., ROIAlign \cite{he2017mask}) directly to a ViT backbone due to the different shapes of the output feature maps of ResNet and ViT. Therefore, we propose a novel Masking with Overlapped Area (MOA) module for feature extraction using a ViT backbone. The MOA module resolves the spatial quantization problem using the overlapped area between each patch and the given region in the attention function. In addition, efficient computation in the MOA module keeps the inference speed similar to that of the ResNet backbone.

To improve the interaction prediction part, we refer to the human perception process of recognizing HOI. Previous studies\cite{gupta2007objects, gupta2009observing, baldassano2017human} show that the HOI perception of humans considers information such as object identity, relative positions, reach motions, manipulation motions, and context. Following this, we assume that HOI recognition by humans consists of three steps (Fig. \ref{fig:human_perception}): 1) Localize each human and object; 2) Identify interactiveness using the spatial relationship between human and object and human pose information; 3) Focus on specific human joints to identify the type of interaction. Motivated by these, we design a pose-conditioned graph neural network for interaction prediction. For identifying the interactiveness part, previous studies exclude non-interactive pairs with auxiliary networks \cite{liu2022interactiveness, li2019transferable}. However, our model represents interactiveness with edge encoding, which is obtained by using the spatial human-object relationship and the human pose information. We can expect more meaningful message passing with interactiveness-aware edge encoding. For focusing on specific joints part, early methods utilize local pose information with simple message passing or attention mechanisms \cite{wan2019pose, zhou2019relation}. 
%Our method updates the human node encodings with the local features of the corresponding human joints by a self-loop structure.
In contrast, we introduce pose-aware attention mechanism using query (spatial features) / key (joint features) structure, resulting in human node encodings containing richer local information by a self-loop structure.

As a result, we propose Vision Transformer based Pose-Conditioned Self-Loop Graph (ViPLO), a novel two-stage HOI detector. We demonstrate the effectiveness of our framework by achieving state-of-the-art results on two public benchmarks: HICO-DET \cite{chao:wacv2018} and V-COCO \cite{gupta2015visual}. In addition, we further evaluate and analyze our model design through extensive ablation studies. The contributions of this paper are as follows: 

% Especially, ViPLO obtains a relative 6.97\% performance gain on the HICO-DET dataset, compared with the very recent state-of-the-art method ParMap \cite{wu2022mining}. 
%We propose a new feature extraction method using Vision Transformer (ViT) backbone. In specific, we introduce a novel \textit{Masking with Overlapped Area} (MOA) module, a suitable feature extraction method for ViT backbone similar to ROIAlign \cite{he2017mask} for ResNet backbone. 
%Also, we design the model structure to be similar to the human perception process using a graphical model. We encode edge encoding using spatial relationship and human pose information (second step in human perception), then update human node encodings with local features of human joints by self-loop structure (third step in human perception). 

\begin{itemize}
    \item {We propose an effective two-stage HOI detector with a ViT backbone by introducing a novel MOA module, which is a suitable feature extraction module for ViT.}
    
    \item {We design the interaction classifier with a pose-conditioned self-loop graph structure, motivated by the human perception process.}
    
    \item {Our model outperforms the previous state-of-the-art results on a widely-used HOI benchmark.}
\end{itemize}

%We want to focus on the second step of this framework. Specifically, MOA module can align .. so that extracted features and regions can be aligned

\begin{figure*}[t]
\begin{center}

\includegraphics[width=0.8\linewidth]{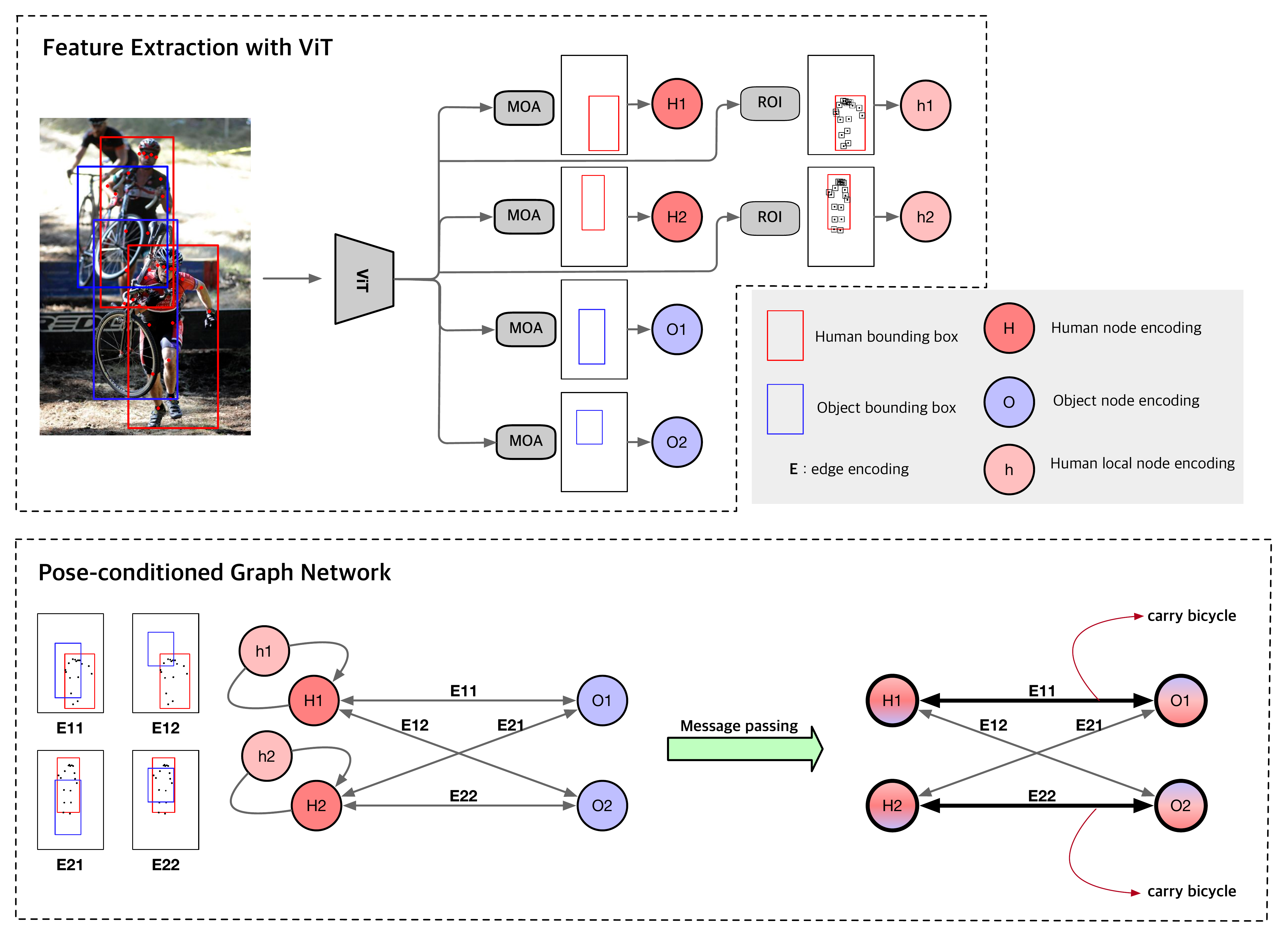}

\end{center}
\vspace{-5mm}
   \caption{Overview of our ViPLO network. We first detect human and objects in a given image with Faster-RCNN \cite{ren2015faster}, then estimate each human pose with an off-the-shelf pose estimator. Then, we extract features for each human and object using a ViT backbone and our novel MOA module. We also extract local features for each human with the estimated pose and ROIAlign \cite{he2017mask}. Next, a graph neural network is used to classify interactions, where the node encoding is initialized with the extracted features and the edge encoding is obtained by spatial and human pose information. Through the message passing process, the human node encoding is also updated with the human local node encoding, helping the model focus on the necessary local part of each human. After the message passing procedure, the combination of human node, object node, and edge encoding becomes the representation of HOI triplet which is used to predict the interaction class.  
}
\vspace{-1mm}
\label{fig:overall}
\end{figure*} 

%------------------------------------------------------------------------
\section{Related Work}
\label{sec:formatting}

\subsection {One-stage Methods}
The one-stage HOI detection framework directly predicts HOI triplets by associating human and objects with predefined anchors, then predicting their interaction. Early methods utilized interaction keypoints \cite{liao2020ppdm, wang2020learning} or union regions \cite{kim2020uniondet} as the predefined anchors. Recently, Transformer-based HOI detectors \cite{kim2021hotr, tamura2021qpic} became the main approach for HOI detection. In particular, HOI Transformers formulated the HOI detection task as a set prediction problem, which are trained with bipartite matching and Hungarian loss. Moreover, unlike the structure of DETR \cite{carion2020end}, methods using two decoders corresponding to each subtask have also been studied \cite{liao2022gen, chen2021reformulating, zhou2022human, zhang2021mining}. However, these methods suffer from low training speed and large memory usage. Besides, since the detection process cannot be separated from the whole framework, it is inefficient in case of already knowing the detection results or wishing to infer interaction only for specific human-object pairs.
\subsection {Two-stage Methods}
The two-stage HOI detection framework detects human and object with an off-the-shelf detector \cite{ren2015faster} and then classifies the interaction label for each human-object pair. 
%Most approaches focused on improving the interaction classifier.
After the appearance of HO-RCNN \cite{chao2018learning}, which is a widely used multi-stream framework, many recent studies use a variety of additional information to get richer contextual features for the interaction classifier, such as spatial features \cite{gao2020drg, zhang2021spatially, ulutan2020vsgnet}, pose features \cite{gupta2019no, li2019transferable, wan2019pose}, and linguistic features \cite{gao2020drg, liu2020consnet}. Several studies \cite{gao2020drg, qi2018learning, ulutan2020vsgnet, zhang2021spatially, wang2020contextual} attempted to encode global contextual information using a message passing mechanism in a graph structure. 
%TIN \cite{li2019transferable} utilized an interactiveness network to exclude non-interactive pairs (Non-Interaction Suppression).

However, there is a lot of room for improvement on the feature extraction step, where existing methods still use ResNet backbones and ROIAlign. % the existing two-stage HOI detection frameworks  do not focus on the feature extr action step, still using the ResNet backbone and ROIAlign.
Since the extracted feature becomes the input of the interaction classifier, the performance of the system would vary significantly depending on the feature extraction method. DEFR \cite{jin2022overlooked} utilized ViT as a backbone network, but considered the HOI recognition task, not HOI detection.
% DEFR also failed to handle the quantization problem, leading to misalignment between the given region and the extracted feature.
DEFR also did not handle the quantization problem, leading to misalignment between the given region and the extracted feature.
%uses the extracted feature only in the inference step, not in the training step. 
Moreover, existing interaction classifiers do not consider the HOI perception process by humans, which focuses on the specific human joints to identify the type of interaction.
%at the end
In contrast, our HOI detector uses a ViT backbone and a graph neural network with pose-conditioned self-loop structure. % which aligns with the HOI perception process by humans.

%------------------------------------------------------------------------
\section{Method}
\label{sec:3}

In this section, we describe our ViPLO in two steps. In Sec. \ref{sec:3-1}, we introduce a novel feature extraction module for the ViT backbone. Then in Sec. \ref{sec:3-2}, we explain our graph neural network with a human pose-conditioned structure for HOI detection. An overview of our framework is shown in Fig. \ref{fig:overall}.

\subsection{Feature Extraction with ViT}
\label{sec:3-1}
We propose a new feature extraction method using the ViT backbone, which enables the following network to recognize interaction effectively. We first detect human and objects in a given image using Faster R-CNN \cite{ren2015faster} (off-the-shelf detector). Then, instead of ResNet, which is a commonly used backbone for feature extraction in conventional two-stage approaches, we use ViT. In the case of the ResNet backbone, visual features are extracted from the feature map of the backbone networks via ROI-Pooling or ROIAlign \cite{he2017mask}. However, for the ViT backbone, a new feature extraction method is needed due to the different shapes of the output feature map compared to the ResNet. Therefore, we introduce a novel \textit{Masking with Overlapped Area} (MOA) module, which is detailed below.
% a suitable feature extraction method for Vision Transformer backbone.

\paragraph{Masking with Overlapped Area}
ViT prepends a learnable embedding (i.e., CLS token) to the sequence of patch embeddings. After passing the transformer encoder, this CLS token serves as an image representation. Therefore to extract a feature corresponding to a given region properly, we have to work with the CLS Token. Inspired by DEFR \cite{jin2022overlooked}, we mask out all patch embeddings outside the given region for the CLS token. For example, to obtain features of an object, the CLS token can only attend to the patch embedding inside the object bounding box. 

However, this approach suffers from the quantization problem as in ROI-Pooling \cite{he2017mask}. Existing ViTs usually use $14 \times 14$, $16 \times 16$, and $32 \times 32$ input patch size. Since the coordinates of the bounding box are given in units of pixels, the edges of the bounding box often do not align with patch boundaries. To solve this problem, we can simply mask out or attend all patch embeddings that the bounding box edge passes through for the CLS token (i.e., quantization). However, this quantization leads to misalignment between the given region and the extracted feature.

\begin{figure}[t]
\begin{center}

\includegraphics[width=0.9\linewidth]{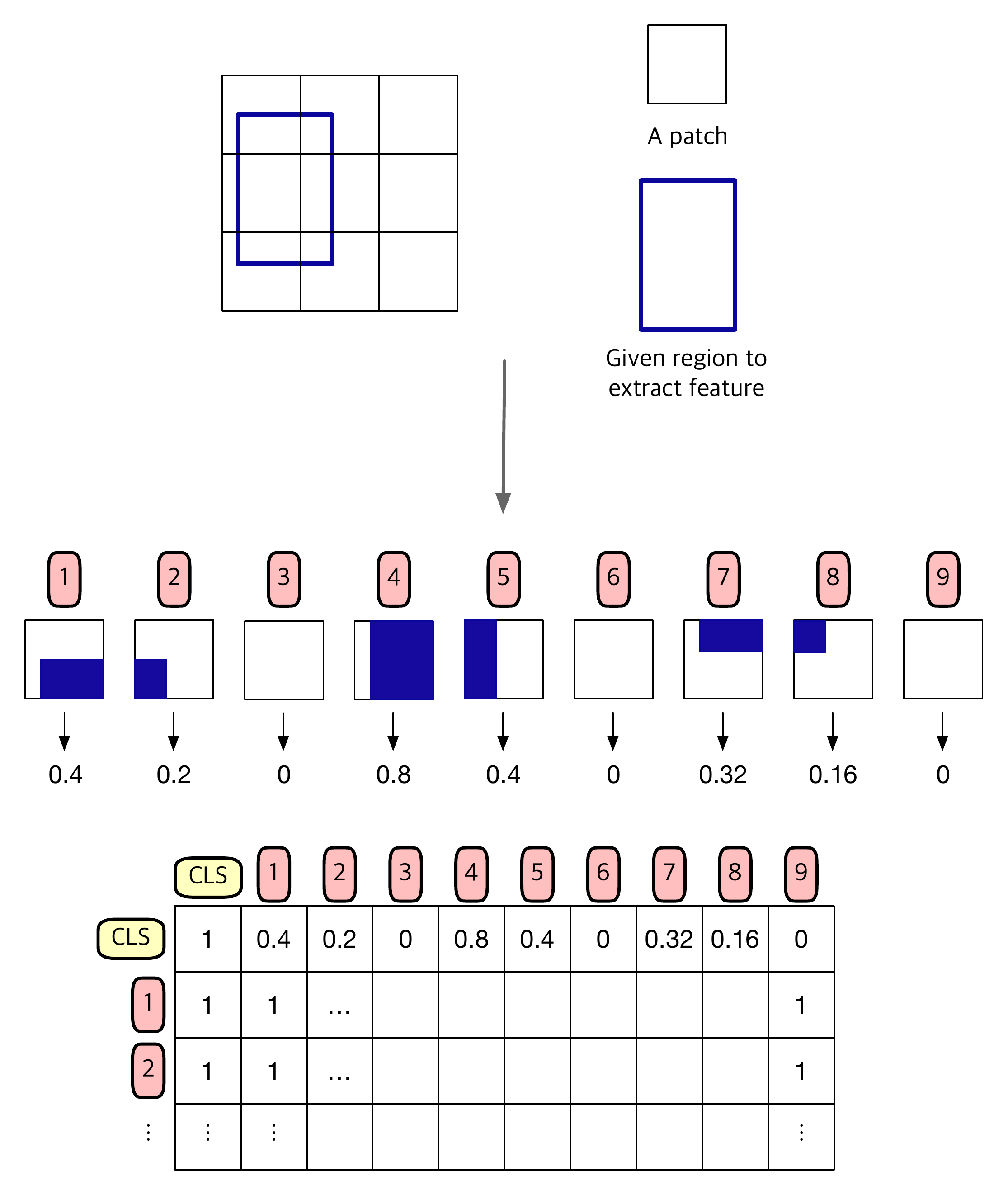}

\end{center}
\vspace{-5mm}
   \caption{Overview of the MOA module. The normalized area (e.g., 0.4, 0.2, 0, ...) between each patch and the given region is calculated. Then, we use these values as the first row of the attention mask matrix (i.e., $S$ in Eq. \ref{eq:1}), whose logarithm is added to the attention map created by the query and key.
}
\vspace{-1mm}
\label{fig:moa}
\end{figure} 

Our MOA module address this quantization problem by utilizing the overlapped area between each patch and the given region to the attention mask in the attention function. We develop our MOA module into a quantization-free layer as shown in Fig. \ref{fig:moa}. First, we calculate the overlapped area of each patch and the given region, and normalize it with the area of the patch. Then, these normalized overlapped area become the first row of the attention mask matrix, which is added to the attention map created by the query and key of the self-attention layer after logarithm.

\begin{align}
    \label{eq:1}
    Attention(Q,K,V) = soft&max(\frac{QK^{T}}{\sqrt{d_{k}}} + \log(S))V,
\end{align}
% Eq. (\ref{eq:1}) demonstrates an attention function in the MOA module applied Transformer.
where $Q$, $K$, and $V$ are the query, key, and value, respectively, and $d_{k}$ is their dimension. $S$ 
%, which is the main part of the MOA module
denotes the normalized overlapped area of the patch and the given region.
% Specifically, $S_{i,j}$ is the normalized overlapped area of the $j$th patch and the given region only when the $i$th token indicates the CLS token. Otherwise, $S_{i,j}$ is set to 1 so that $\log(S)$ does not affect the attention operation. In this way, the CLS token cannot attend to a patch outside the given region, because the normalized overlapped area is 0. For the patch which the bounding box edge passes through, the attention score is adjusted in proportion to the overlapped area.

\paragraph {Efficient computation for MOA}
% computation? implementation?
The MOA module leads to a large performance increment in HOI detection, as shown in the ablation studies in Sec. \ref{sec:4-3}. To use the MOA module, however, overlapped area $S$ has to be computed for each bounding box, which may be a computational burden. We design the entire process of computing $S$ to be run through GPU operations, and apply techniques for reducing computations. See Appendix \ref{app:b} for more details about computational complexity and implementation of the MOA module.

\subsection{Pose-conditioned Graph Neural Network}
\label{sec:3-2}
After extracting features with our MOA module, we use a graph neural network to detect interactions between human and objects. Inspired by the process of HOI detection by humans, we improve the spatially conditioned graph (SCG) \cite{zhang2021spatially} by using human pose information. We first review the baseline SCG, and then describe our improved network design.

\paragraph {SCG Baseline}
SCG is a bipartite graph neural network with human and object nodes. The features extracted using ResNet and ROIAlign are used to initialize the node encodings. The edge encodings are initialized with features based on the spatial information of two bounding boxes, i.e., human and object. Then, bidirectional message passing is performed between nodes conditioned on their edge encodings. After message passing, the updated node encodings and edge encodings are used to classify interaction between the human and object. 

% \paragraph{Pose-aware Edge encoding}
%  As mentioned in the introduction, the second step of human HOI perception process is identifying interactiveness using a spatial information between human and objects and human pose information. In the graph structure, since the relationship between human nodes and object nodes is expressed by edge encoding, we initialize edge encoding based on human pose information as well as spatial information. In particular, we compute pairwise spatial features (i.e., query) same as SCG. Then, we compute the pairwise joint features (i.e., key) by passing MLP to the handcrafted joint features, including each joint coordinate and directional vector from each joint coordinate to object box center coordinates. Following the attention mechanism, we compute the attention score for each human joint by dot product of query and key as follows:
 
 \paragraph{Pose-aware Edge encoding}
 As mentioned in the introduction, the second step of the HOI perception process is identifying interactiveness using the spatial information between human and objects and human pose information. In the graph structure, the relationship between a human node and object node is expressed by the edge encoding between the two nodes. Thus, we initialize the edge encoding based on human pose information as well as spatial information. In particular, we compute pairwise spatial features (i.e., query) as in SCG. Then, we compute the pairwise joint features (i.e., key) by applying MLP to the handcrafted joint features including the coordinates of each joint and the directional vector from the joint to the object box center. Following the attention mechanism, we compute the attention score for each human joint by the dot product of the query and key as follows:

\begin{align}
    \label{eq:4}
    \boldsymbol{\alpha}_{ij} &= softmax(Q_{ij}K_{ij}^{T} \cdot s_{i}),
\end{align}
where $\boldsymbol{\alpha}_{ij}$, $Q_{ij}$, and $K_{ij}$ are the joint attention weight, pairwise spatial feature, and pairwise joint feature for the $i$th human and $j$th object, respectively, and $s_{i}$ is the pose estimation confidence score for the $i$th human. The obtained attention score is used in the following message passing step. The edge encoding is initialized with a pairwise spatial feature, but becoming pose-aware due to the attention mechansim. 
%Since human pose information is crucial for the relationship between human and objects, we initialize edge encoding based on human pose information as well as spatial information. Specifically, we compute features with pose information by including each joint coordinate and directional vector from each joint coordinate to object box center coordinates. These feature is passed through MLP to output an edge encoding. 

\paragraph {Message Passing with Pose}
As described in \cite{wan2019pose}, there are cases where detailed local features are required for recognizing interaction (e.g., catch and grab). This is related to the third step of the HOI perception process by humans, focusing on specific human joints to identify the interaction. The pose-aware edge encoding alone cannot adjust the model to utilize local information. In our method, we incorporate the local information in the human node encoding, i.e., we update the human node encoding using the local features of each human joint by new message functions. 

First, we extract the local features by applying ROIAlign to the ViT output for the local region box for each human joint, which is inspired by \cite{wan2019pose}. Then, we compute the weighted sum of each local feature of a human joint to get the human local feature:
\begin{align}
    %\label{eq:4}
    %\boldsymbol{\alpha}_{ij} &= \textrm{MLP}(\mathbf{z}_{ij}), \\
    \label{eq:5}
    \mathbf{x}_{ij,local} &= \sum_{k}{\alpha_{ijk} \odot \mathbf{x}_{ik,local}},
\end{align}
where $\alpha_{ijk}$ is the $k$th value of joint attention weight $\boldsymbol{\alpha}_{ij}$, $\mathbf{x}_{ik, local}$ is the local feature for the $k$th joint of the $i$th human, and $\mathbf{x}^{t}_{ij,local}$ is the $i$th human local feature for the $j$th object. $\odot$ denotes the element-wise multiplication.

To use ROIAlign with the ViT output, we reshape the image patch embeddings to the form of a square feature map, i.e., Unflatten. After the unflatten process, the reshaped patch embeddings play the same role as ResNet output feature maps, so we can simply apply ROIAlign to the patch embeddings. %We choose ROIAlign rather than the MOA module for extracting local features from VIT output features, even though the token is not used in ROIAlign. 
We hypothesize that the image patch tokens contain more detailed information rather than the CLS token, choosing the ROIAlign to extract local features rather than the MOA module. When training ViT, the classification head is attached to the final CLS token output. Therefore, the CLS token has coarse information for image classification. On the other hand, patch tokens, which become the components of CLS tokens (i.e., attention mechanism), have fine information for each region. This hypothesis is supported by the ablation study in Sec. \ref{sec:4-3}.
%and visualized result in Sec. \ref{sec:4-4}.

Second, the extracted human local features ($\mathbf{x}_{ij,local}$) become human local node encodings, which are used to update the human node encodings (Fig. \ref{fig:overall}). In detail, the human node encodings are iteratively updated with a message from the human local features, as well as object node encodings and edge encodings:

\begin{align}
    \label{eq:2}
    M_{\mathcal{O} \rightarrow \mathcal{H}}(\mathbf{x}^{t}_{i}, \mathbf{y}^{t}_{j}, \mathbf{z}_{ij}) &=  \textrm{MBF}_{o}(\mathbf{x}_{ij,local} \oplus \mathbf{y}^{t}_{j},\mathbf{z}_{ij}), \\
    \label{eq:3}
    M_{\mathcal{H} \rightarrow \mathcal{O}}(\mathbf{x}^{t}_{i}, \mathbf{z}_{ij}) &=  \textrm{MBF}_{h}(\mathbf{x}^{t}_{i},\mathbf{z}_{ij}),
\end{align}
where $M_{\mathcal{O} \rightarrow \mathcal{H}}$ denotes the message function from object nodes to human nodes, and vice versa. $\mathbf{x}^{t}_{i}$ and $\mathbf{y}^{t}_{j}$ are the $i$th human node encoding and  $j$th object node encoding, respectively, at message passing step $t$. $\textrm{MBF}$ indicates the multi-branch fusion module proposed in \cite{zhang2021spatially}. $\oplus$ denotes the concatenation operation. %To avoid confusion, the notations in this subsection are the same as the baseline model, SCG.
% This algorithm is similar to self-loop in graph theory, that human node encodings are updated with the human local features similar to itself.
The human nodes are updated not only with the object nodes but also with the human local features which share the similar information with the human node encodings. This algorithm resembles the self-loop structure in graph theory.

\begin{figure}[t]
\begin{center}

\includegraphics[width=.4\linewidth]{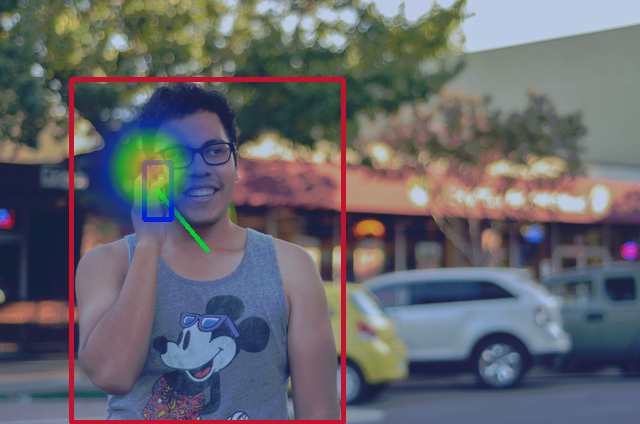}
\includegraphics[width=.4\linewidth]{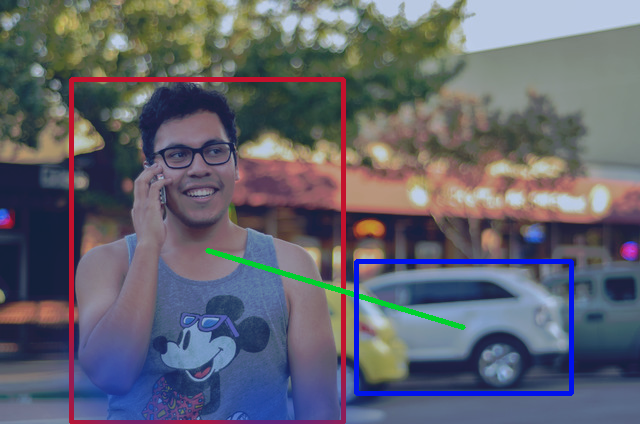}

\end{center}
\vspace{-5mm}
  \caption{Heatmaps showing joint attention. \textbf{Left}: ViPLO focuses on the local information of hand to detect interaction between cellphone and human. \textbf{Right}:  ViPLO does not focus on a specific joint for local information when detecting interaction between car and human, resulting in no-interaction.}
\vspace{-1mm}
\label{fig:qualitative}
\end{figure}

\begin{table*}[t]
\begin{center}
\resizebox{1.\linewidth}{!}{
\begin{tabular}{llcccccccccc}
\hline
& & \multicolumn{7}{c}{\textbf{HICO-DET}} && \multicolumn{2}{c}{\textbf{V-COCO}} \\
\multirow{3}{*}{Method} & \multirow{3}{*}{Backbone} & \multicolumn{3}{c}{Default} && \multicolumn{3}{c}{Known Object} \\
\cmidrule{3-5} \cmidrule{7-9} \cmidrule{11-12}
&& Full & Rare & Non-Rare && Full & Rare & Non-Rare && $AP^{S1}_{role}$ & $AP^{S2}_{role}$  \\
\hline\hline
\multicolumn{5}{l}{\emph{One-stage methods}} \\
UnionDet \cite{kim2020uniondet} & ResNet-50-FPN & 17.58 & 11.72 & 19.33 && 19.76 & 14.68 & 21.27 && 47.5 & 56.2 \\
IP-Net \cite{wang2020learning}   & Hourglass-104 & 19.56 & 12.79 & 21.58 && 22.05 & 15.77 & 23.92 && 51.0 & - \\
PPDM \cite{liao2020ppdm} & Hourglass-104 & 21.73 & 13.78 & 21.40 && 24.58 & 16.65 & 26.84 && - & -  \\
GG-Net \cite{zhong2021glance}  & Hourglass-104 & 23.47 & 16.48 & 25.60 && 27.36 & 20.23 & 39.48 && 54.7 & -\\
HOTR \cite{kim2021hotr} & ResNet-50 &25.10 & 17.34 & 27.42 && - & - & - && 55.2 & 64.4 \\
HOI-Trans \cite{zou2021end}   & ResNet-101 & 26.61 & 19.15 & 28.84 && 29.13 & 20.98 & 31.57 && 52.9 & - \\
AS-Net \cite{chen2021reformulating} & ResNet-50 & 28.87 & 24.25 & 30.25 && 31.74 & 27.07 & 33.14 && 53.9 & - \\
QPIC \cite{tamura2021qpic}  & ResNet-101 & 29.90 & 23.92 & 31.69 && 32.38 & 26.06 & 34.27 && 58.3 & 60.7 \\
MSTR \cite{kim2022mstr} & ResNet-50 & 31.17 & 25.31 & 32.92 && 34.02 & 28.83 & 35.57 && 62.0 & 65.2 \\
SSRT \cite{iftekhar2022look}  & ResNet-101 & 31.34 & 24.31 & 33.32 && - & - & - && \underline{65.0} & 67.1 \\
CDN \cite{zhang2021mining}  & ResNet-101 & 32.07 & 27.19 & 33.53 && 34.79 & 29.48 & 36.38 && 63.9 & 65.9 \\
DOQ \cite{qu2022distillation} & ResNet-50 & 33.28 & 29.19 & 34.50 && - & - & - && 63.5 & - \\
IF \cite{liu2022interactiveness} & ResNet-50 & 33.51 & 30.30 & 34.46 && 36.28 & 33.16 & 37.21 && 63.0 & 65.2\\
GEN-VLKT \cite{liao2022gen} & ResNet-101 & 34.95 & 31.18 & \underline{36.08} && \underline{38.22} & 34.36 & \underline{39.37} && 63.6 & 65.9 \\
ParMap \cite{wu2022mining} & ResNet-50 & \underline{35.15} & 33.71 & 35.58 && 37.56 & 35.87 & 38.06 && 63.0 & 65.1\\
\hline
\multicolumn{5}{l}{\emph{Two-stage methods}} \\
VCL \cite{hou2020visual}    & ResNet-50 & 23.63 & 17.21 & 25.55 && 25.98 & 19.12 & 28.03 && 48.3 & -\\
ATL \cite{hou2021affordance}  & ResNet-50 & 23.67 & 17.64 & 25.47 && 26.01 & 19.60 & 27.93 && - & -\\
IDN \cite{li2020hoi}  & ResNet-50 & 24.58 & 20.33 & 25.86 && 27.89 & 23.64 & 29.16 && 53.3 & 60.3\\
FCL \cite{hou2021detecting} & ResNet-50 & 24.68 & 20.03 & 26.07 && 26.80 & 21.61 & 28.35 && 52.4 & - \\
SCG \cite{zhang2021spatially} & ResNet-50-FPN & 29.26 & 24.61 & 30.65 && 32.87 & 27.89 & 34.35 && 54.2 & 60.9 \\
STIP \cite{zhang2022exploring} & ResNet-50 & 32.22 & 28.15 & 33.43 && 35.29 & 31.43 & 36.45 && \textbf{66.0} & \textbf{70.7} \\
DEFR \cite{jin2022overlooked} & ViT-B/16 & 32.35 & 33.45 & 32.02 && - & - & - && - & - \\
UPT \cite{zhang2022efficient} & ResNet-101 & 32.62 & 28.62 & 33.81 && 36.08 & 31.41 & 37.47 && 61.3 & 67.1 \\
\rowcolor{Gray}
$\textrm{ViPLO}_{s}$ & ViT-B/32 & 34.95 & \underline{33.83} & 35.28 && 38.15 & \underline{36.77} & 38.56 && 60.9 & 66.6 \\
\rowcolor{Gray}
$\textrm{ViPLO}_{l}$ & ViT-B/16 & \textbf{37.22} & \textbf{35.45} & \textbf{37.75} && \textbf{40.61} & \textbf{38.82} & \textbf{41.15} && 62.2 & \underline{68.0} \\
\hline
\end{tabular}}
\end{center}
\vspace{-0.2in}
\caption{Performance comparison in terms of mAP on the HICO-DET and V-COCO datasets. For fair comparison between one-stage methods and two-stage methods, we report results using an object detector finetuned on the training dataset for two-stage methods. We use the finetuned object detector from \cite{zhang2022efficient}, i.e., DETR with the ResNet-101 backbone. We do not consider the results using detections from \cite{gao2020drg} due to a problem of reproductibility. In each evaluation setting, the best result is marked with bold and the second best result is underlined.} 
\label{tab:compare-all}
\end{table*}

\paragraph{Effectiveness of Pose-Conditioned Graph}
Our pose-aware edge encoding scheme helps models focus only on the necessary message passing. For example, the case where a human is just standing next to the object and the case where a hand is extended to the direction of the object are different, even though the spatial relationship of the region pair is the same. We can expect more meaningful message passing in the latter case, due to the different edge encoding from each human pose information. Moreover, human local features can contain richer local information by the attention operation in Eq. \ref{eq:5} (hand in the latter case). Furthermore, as shown in Fig. \ref{fig:qualitative}, human node encodings are updated with self-loop architecture, to focus only on the necessary local part as the message passing step progresses. These human node encodings with rich local information also enrich the object node encodings with human-to-object messages, resulting in effective HOI detection. 

\subsection{Training and Inference}
\label{sec:3-3}
We follow the training and inference procedure of the SCG \cite{zhang2021spatially, zhang2022efficient}. From the pose-conditioned graph after the message passing, we get interaction classification scores for each pair of human and object node using the corresponding edge encoding. The focal loss \cite{lin2017focal} is used as the multi-label classification loss to train the possible interactions for each human-object pair.

\section{Experiments}
\label{sec:4}
In this section, we first introduce our experimental settings (Sec. \ref{sec:4-1}). We then compare our proposed model with state-of-the-art approaches (Sec. \ref{sec:4-2}). Finally we conduct ablation studies (Sec. \ref{sec:4-3}) demonstrating the advantages and effectiveness of our model. Qualitative results can be found in Appendix \ref{app:c}.

\subsection{Experimental Settings}
\label{sec:4-1}
\paragraph{Datasets}
We evaluate our model on two public datasets, HICO-DET \cite{chao:wacv2018} and V-COCO \cite{gupta2015visual}, which are widely used in HOI detection. HICO-DET contains 47,776 images (38,118 training images and 9,658 test images), with 80 object categories (same categories with the MS-COCO \cite{lin2014microsoft} dataset) and 117 verb classes constructing 600 types of HOI triplets. HICO-DET provides more than 150k annotated human-object pairs. V-COCO is a subset of MS-COCO, which is a much smaller dataset than HICO-DET. It contains 10,346 images (2,533 training images, 2,867 validation images, and 4,946 test images), with the same 80 object categories and 29 verb classes. Since HICO-DET containing various verb types is more suitable in the real world, we focus our evaluation mainly on the HICO-DET dataset.

\paragraph{Evaluation Metrics}
We follow the standard settings in \cite{gao2018ican}, reporting mean Average Precision (mAP) for evaluation. Prediction of a HOI triplet is considered as a true positive when both predicted human and object bounding boxes have IoUs larger than 0.5 compared to the ground truth boxes, and HOI category prediction is accurate. For HICO-DET, we report mAP over two evaluation settings (Default and Known Object), with three HOI category subsets: all 600 HOI triplets (Full), 138 HOI triplets with fewer than 10 training samples (Rare), and 462 HOI triplets with 10 or more training samples (Non-Rare). For V-COCO, we report mAP for two scenarios: including a prediction of the occluded object bounding box (Scenario 1), or ignoring such a prediction (Scenario 2). 

\paragraph{Implementation Details}
%We change the model architecture from SCG (i.e., feature extractor and graph neural network), while maintaining the training, inference, and generating region proposals processes.
Since neither the HOI Detection dataset nor the Faster R-CNN detector result contains a ground truth or estimated human pose, we need an off-the-shelf pose estimator. We apply Vitpose \cite{xu2022vitpose} as a pose estimator, which is trained on the MS-COCO Keypoint dataset \cite{lin2014microsoft}. As a result, each human bounding box (including the ground truth human box) has 17 keypoints. 

We consider the following two variants of our method: 1) $\textrm{ViPLO}_{s}$: a small version of ViPLO, which uses ViT-B/32 as the backbone; and 2) $\textrm{ViPLO}_{l}$: large version of ViPLO, which uses ViT-B/16 as the backbone. All other hyperparameters are the same between the two models. The feature extractor (i.e., backbone) is initialized using CLIP \cite{radford2021learning} with pre-trained weights. We cannot directly apply CLIP image pre-processing, due to the center-crop process which is not suitable for object detection. Hence, we resize input images to $672 \times 672$ pixels using data transformation in \cite{huang2020devil}. Bounding boxes and human joints are resized accordingly. Vit-B is pre-trained on $224 \times 224$ pixels, but the model can be fine-tuned on $672 \times 672$ pixels with 2D interpolation of the pre-trained position embeddings, as described in \cite{dosovitskiy2020image}. Further implementation detail can be found in Appendix \ref{app:a}.

\begin{table}[t]
\begin{center}
\resizebox{1.\linewidth}{!}{
\begin{tabular}{llccc}
\hline
\multirow{3}{*}{Method} & \multirow{3}{*}{Backbone} & \multicolumn{3}{c}{Default} \\
\cmidrule{3-5} && Full & Rare & Non-Rare  \\
\hline\hline
\multicolumn{5}{l}{\emph{Two-stage methods with ground truth boxes}} \\
iCAN \cite{gao2018ican} & ResNet-50 & 33.38 & 21.43 & 36.95 \\
TIN \cite{li2019transferable}  & ResNet-50 & 34.26 & 22.90 & 37.65 \\
VCL \cite{hou2020visual}  & ResNet-50 & 43.09 & 32.56 & 46.24 \\
IDN \cite{li2020hoi}  & ResNet-50 & 43.98 & 40.27 & 45.09 \\
FCL \cite{hou2021detecting} & ResNet-50 & 44.26 & 35.46 & 46.88 \\
ATL \cite{hou2021affordance} & ResNet-50 & 44.27 & 35.52 & 46.89 \\
SCG \cite{zhang2021spatially} & ResNet-50-FPN & 51.53 & 41.01 & 54.67 \\
\rowcolor{Gray}
$\textrm{ViPLO}_{s}$ & ViT-B/32 & \underline{58.23} & \underline{54.08} & \underline{59.46} \\
\rowcolor{Gray}
$\textrm{ViPLO}_{l}$ & ViT-B/16 & \textbf{62.09} & \textbf{59.26} & \textbf{62.93} \\
\hline
\end{tabular}}
\end{center}
\vspace{-0.2in}
\caption{Performance comparison in terms of mAP on the HICO-DET dataset with ground truth detection results. The best result is marked in bold and the second best result is underlined.}
\label{tab:hicodet-res-gt}
\end{table}

\begin{table}[t]
\begin{center}
\resizebox{1.\linewidth}{!}{
\begin{tabular}{l|l|ccc}
\hline
\multirow{2}{*}{Method} & \multirow{2}{*}{Backbone} &\multirow{2}{*}{mAP $\uparrow$} &\multirow{2}{*}{GPU memory $\downarrow$} & \multirow{2}{*}{Speed $\downarrow$} \\
& & & & \\
\hline\hline
SCG \cite{zhang2021spatially} & ResNet-50 & 29.26 & 3423 MiB & 106 ms\\
$\textrm{ViPLO}_{s}$ & ViT-B/32 & 33.92 & 2531 MiB & 110 ms\\
$\textrm{ViPLO}_{l}$ & ViT-B/16 & 36.97 & 3119 MiB & 131 ms\\
\hline
\end{tabular}}
\end{center}
\vspace{-0.2in}
\caption{Comparison of mAP, memory usage, and speed for inference on the HICO-DET test dataset. We exclude the pose-conditioned graph part in our model for a fair comparison. We conduct an inference process with a batch size of 1 on a Geforce RTX 3090 GPU. ``Speed" column means the elapsed time for processing one image.}
\label{tab:speed-memory}
\end{table}

\begin{table*}[ht!]\centering
\captionsetup[subtable]{font=small}
\begin{subtable}{0.4\textwidth}
{\begin{tabular*}{\linewidth}{l|ccc}
 Methods & Full & Rare & Non-Rare \\
\hline\hline
\rule{0pt}{0.9\normalbaselineskip}SCG (ResNet-50) & 29.26 & 24.61 & 30.65\\
ViT + ROI & 32.84 & 28.48 & 34.14 \\
ViT + $\textrm{MOA}^{Q}$ &  33.20 & \textbf{31.26} & 33.78  \\
ViT + MOA &  \textbf{33.92} & 30.84 & \textbf{34.83} \\
\hline
\end{tabular*}} 
\caption{Effect of ViT backbone and MOA module.}
\label{tab:abl-moa}
\end{subtable} \hspace{3mm}
\begin{subtable}{0.55\textwidth} 
{\begin{tabular*}{\linewidth}{l|ccc}
  Methods & Full & Rare & Non-Rare\\

\hline\hline
\rule{0pt}{0.9\normalbaselineskip}Base (ViT + MOA) & 33.92 & 30.84 & 34.83\\
+ pose edge & 34.51 & 32.74 & 35.04 \\
+ pose edge + local pose (MOA) & 34.44 & 31.21 & \textbf{35.41} \\
+ pose edge + local pose (ROI) & \textbf{34.95} & \textbf{33.83} & 35.28 \\
\hline
\end{tabular*}}
\caption{Effect of the pose-conditioned graph neural network.}
\label{tab:abl-graph}
\end{subtable}
\caption{Ablation study of ViPLO components on the HICO-DET test dataset under the Default setting.}
\label{tab:ablation}
\end{table*}

\subsection{Comparison to State-of-the-Art}
\label{sec:4-2}
Table \ref{tab:compare-all} shows the performance comparison of ViPLO and other state-of-the-art methods, which are grouped into one-stage and two-stage methods. For HICO-DET, our $\textrm{ViPLO}$ outperforms all existing one-stage and two-stage methods by a large margin in each evaluation setting. Specifically, $\textrm{ViPLO}_{l}$ achieves new state-of-the-art performance of \textbf{37.22} mAP in the Default Full setting, obtaining a relative performance gain of 5.89\% compared to the most recent ParMap \cite{wu2022mining}. Compared with UPT \cite{zhang2022efficient}, which is the previous state-of-the-art two-stage method, our model achieves a significant relative performance gain of 14.10\%. These results indicate the effectiveness of our method, utilizing the ViT backbone with the MOA module, and updating human features with the pose-conditioned self-loop structure. For V-COCO, our $\textrm{ViPLO}$ achieves comparable performance to previous state-of-the-art methods. The performance gain is not as noticeable as on HICO-DET, due to the fewer training samples which can be fatal to the ViT architecture \cite{touvron2021training}.

As mentioned in the introduction, an advantage of two-stage methods is that
% ground truth detection results can be used for recognizing human-object interaction.
they can also be used for HOI recognition when ground truth bounding boxes are available. Performance comparison using ground truth boxes for HICO-DET is shown in Table \ref{tab:hicodet-res-gt}. Our method outperforms all previous state-of-the-art methods with a significant large margin of 10.56 in terms of mAP in the Default Full setting. 

Meanwhile, since most previous methods use ResNet as the backbone network, memory efficiency and speed of our method using the ViT backbone are examined. As shown in Table \ref{tab:speed-memory}, we obtain a large performance improvement in terms of mAP, while maintaining speed and memory efficiency at a similar level to the ResNet-50 backbone. Surprisingly, $\textrm{ViPLO}$ consumes less GPU memory than SCG using ResNet-50 as the backbone network. These results show the efficiency and efficacy of the ViT backbone and proposed MOA module.

\subsection{Ablation Study}
\label{sec:4-3}
In this subsection, we conduct ablation studies to evaluate and analyze the effectiveness of our proposed module and our model design. All experiments are conducted on the HICO-DET test dataset under the Default setting, with a small version of VIPLO, i.e., $\textrm{ViPLO}_{s}$.

\paragraph{MOA module} Our model adopts the ViT backbone and MOA module for feature extraction, where this feature serves as the node encoding in the subsequent graph neural network. We explore the benefits of the ViT backbone and MOA module, as shown in Table \ref{tab:abl-moa}. We exclude the pose-conditioned graph structure for a fair comparison with the SCG baseline. ViT + ROI (Row 2) means the case extracting features with reshaped patch embeddings and ROIAlign for the bounding box, which is described in the human local feature extraction process in Sec. \ref{sec:3-2}. ViT + $\textrm{MOA}^{Q}$ (Row 3) means the case using the MOA module without calculating the overlapped area, and simply masking out the patch embeddings that bounding box edges pass through (i.e., quantization). Compare to the SCG baseline, the ViT backbone significantly improves the performance by 3.58 in terms of mAP for the Full setting, even without the MOA module. Further using our MOA module (ViT + MOA) yields additional performance improvement by 1.08 in mAP, demonstrating the effectiveness of the MOA module. In particular, the performance difference depending on the quantization of the MOA module supports the importance of the quantization-free method by calculating the overlapped area, which is a key part of our MOA module.

\paragraph{Pose-conditioned Graph Neural Network}
Table \ref{tab:abl-graph} explore the effectiveness of the components of pose-conditioned graph neural network, on top of the ViT backbone and MOA module (Row 1). We first add the human pose information to formulate the edge encoding (Row 2), which shows a performance improvement of 0.59 in mAP. Moreover, adding a self-loop structure with human local node encodings boosts the performance by 0.44 in mAP, proving the effectiveness of a self-loop structure that aligns with the perception process by humans (Row 4). However, extracting human local features by the MOA module does not boost the performance, it actually decreases the performance (Row 3). %This result supports our hypothesis in Sec. \ref{sec:3-2}. 
%In fact, there are many images in the HICO-DET dataset that do not reveal a human's whole body, which makes pose estimation difficult (e.g., an image with only a hand and object). We observe more performance gaps between Table \ref{tab:abl-graph} methods when testing on the partial HICO-DET test dataset, only including images with high pose estimation scores. Further details and results partial HICO-DET test dataset can be found in Appendix. 

\begin{table}[t]
\begin{center}
\resizebox{1.\linewidth}{!}{
\begin{tabular}{l|ccc}
\hline
\multirow{2}{*}{Method} & \multirow{2}{*}{Backbone} & \multirow{2}{*}{Pre-trained Dataset} &\multirow{2}{*}{mAP $\uparrow$}\\
& & & \\
\hline\hline
SCG & ResNet50 & MS COCO &29.26 \\
ViPLO & ResNet50 & CLIP & 32.23 \\
$\textrm{ViPLO}_{s}$ & ViT-B/32 & CLIP &33.92 \\
$\textrm{ViPLO}_{l}$ & ViT-B/16 & CLIP &36.97 \\
\hline
\end{tabular}}
\end{center}
\vspace{-0.2in}
\caption{Effect of CLIP pre-trained parameters. Experiment setting is the same as in Table \ref{tab:speed-memory}}
\label{tab:clip-parm}
\end{table}

\paragraph {CLIP pre-trained parameters}
As we mentioned, the feature extractor (i.e., ViT backbone) is initialized with CLIP pre-trained mode. However, the ResNet backbone of SCG is initialized with COCO pre-trained parameters, which is trained with much less amount of images than CLIP. Therefore, we conduct an additional experiment comparing the ResNet backbone and ViT backbones with both the same CLIP pre-trained parameters, as shown in Table \ref{tab:clip-parm}. Please note that we cannot apply the same preprocess/feature extraction method for Row 1 and Row 2, due to the modified architecture of CLIP ResNet50, such as the attention pooling mechanism. So we adopt the preprocess/feature extraction method of Row 2 in the same way
as ViT + ROI in Table \ref{tab:abl-moa} in the paper. We can still observe the superior performance of our model in the same pre-trained mode, showing the effectiveness of our ViT + MOA scheme. 

%\begin{figure}[t]
%\begin{center}

%\includegraphics[width=1.\linewidth]{qualitative.pdf}

%\end{center}
%\vspace{-5mm}
%  \caption{Qualitative results of $\textrm{ViPLO}_{l}$ compared to the baseline SCG. For each image, the prediction scores of $\textrm{ViPLO}_{l}$ and SCG are shown (left: $\textrm{ViPLO}_{l}$, right: SCG). The joint attention in Eq. \ref{eq:4} is also visualized as a heatmap.}
%\vspace{-1mm}
%\label{fig:qualitative}
%\end{figure} 

%\subsection{Visualization Results}
%\label{sec:4-4}
%We show qualitative results of ViPLO compared to the SCG in Fig. \ref{fig:qualitative}. We find that ViPLO can successfully detect difficult interactions where the human and object are far away (case 1). ViPLO also effectively detects interactions focusing on specific human joint, such as ankle or wrist in case of riding skateboard (case 2). Surprisingly, we find that our model focuses on different joints when detecting interaction for different objects, even in the same image (case 3). These results prove the effectiveness of ViPLO. 
%pose 이야기..? 

\section{Conclusion}
\label{sec:5}

In this paper, we have proposed the ViPLO, the state-of-the-art two-stage HOI detector assisted by the MOA module and pose-conditioned graph. With the MOA module, our detector fully utilizes the ViT as the feature extractor by addressing the quantization problem. In addition, the pose-conditioned graph inspired by the HOI perception process of humans makes our detector exploit the rich information from human poses. In addition, unlike the one-stage methods, ViPLO has the advantage of low complexity and convenience to apply in real world scenarios.

%%%%%%%%% REFERENCES
{\small
\bibliographystyle{ieee_fullname}
\bibliography{egbib}
}

\clearpage

\appendix

\section{Implementation detail}
\label{app:a}
We implement our model ViPLO on top of the Pytorch implementation \footnote{\url{https://github.com/fredzzhang/spatially-conditioned-graphs}} of SCG\cite{zhang2021spatially} . We also use the Pytorch implementation \footnote{\url{https://github.com/openai/CLIP}} of CLIP \cite{radford2021learning} for the backbone network, including modifications related to the proposed MOA module. As mentioned in Sec. 3.3, we follow the training and inference procedure of SCG, such as appending the ground-truth boxes during the training, applying non-maximum suppression (NMS) to the detection results, and computing the final HOI scores for the focal loss. The final HOI scores are computed as in SCG: 
% We implement our model ViPLO on top of the SCG \cite{zhang2021spatially} Pytorch implementation \footnote{\url{https://github.com/fredzzhang/spatially-conditioned-graphs}}. We also use the CLIP \cite{radford2021learning} Pytorch implementation \footnote{\url{https://github.com/openai/CLIP}} for the backbone network, including modifications related to the proposed MOA module. As said in the Sec. 3.3, we follow the training and inference procedure of the SCG, such as appending the ground-truth boxes during training, applying non-maximum suppression(NMS) to the detection results, and computing the final HOI scores for the focal loss. The final HOI scores are computed following SCG : 
\begin{align}
    \label{eq:1}
    \mathbf{s}_{k} = (s^{h}_{i})^{\lambda} \cdot (s^{o}_{j})^{\lambda} \cdot \Tilde{s_{k}}, 
\end{align}
where $s^{h}_{i}$ denotes the $i$th human detection score, $s^{o}_{j}$ denotes the $j$th object detection score, and $\Tilde{s_{k}}$ is the action classification score obtained from the representation of the HOI triplet, including human node encoding, object node encoding, and their edge encoding. These encodings are fused with the MBF module in the SCG. We set $\lambda$ to 1 in the training process and 2.8 in the inference process \cite{zhang2021spatially}.
% We follow the training and inference procedure of the SCG \cite{zhang2021spatially, zhang2022efficient}. From the pose-conditioned graph after the message passing, we get interaction classification scores for each pair of human and object node using the corresponding edge encoding.
Finally, the focal loss \cite{lin2017focal} is used as the multi-label classification loss to train the possible interactions for each human-object pair as follows.
\begin{align}
    \label{eq:6}
    FL(\hat{y}, y) = \begin{cases}
      -\alpha(1-\hat{\mathit{y}})^{\gamma}\log(\hat{\mathit{y}}), & y = 1\\
      -(1-\alpha)\hat{\mathit{y}}^{\gamma}\log(1-\hat{\mathit{y}}), & y = 0
    \end{cases}
\end{align}
where $\mathit{y}$ is the ground-truth label, $\hat{\mathit{y}}$ is the final score for the human-object pair, and $\alpha$ and $\gamma$ are balancing parameters.
For focal loss, we set $\alpha$ to 0.5 and $\gamma$ to 0.2 \cite{zhang2021spatially}. 

% When using the Vision Transformer backbone, CLS tokens in which the MOA module is applied are mapped to initialization of each node encoding with two-layer MLP. We use three-layer MLP to output an edge encoding from human pose and spatial information. For extracting local feature of human, we draw the local region box for each joint as 0.3 times the size of the human box height. For the message function using human local node encodings and object node encodings (Eq. 4 in the paper), we concat two node encodings for appearance feature in the MBF module. 

When using the Vision Transformer backbone, CLS tokens in which the MOA module is applied are mapped to initialization of each node encoding with a two-layer MLP. We use a three-layer MLP to construct an edge encoding from human pose and spatial information. For extracting the local feature of human, we draw the local region box for each joint as 0.3 times the size of the human bbox height. For the message function using human local node encodings and object node encodings (Eq. 4 in the paper), we concatenate two node encodings for the appearance feature in the MBF module. 

% We use AdamW \cite{loshchilov2017decoupled} optimizer for training with initial learning rate of $10^{-4}$. For HICO-DET, we train the VIPLO for 8 epochs with a learning rate decay by the factor of 0.1 at the $6^{th}$ epochs with flip data augmentation. For V-COCO, we train the model for 20 epochs with 0.1 times the learning rate decay at the $10^{th}$ epochs, with additional data augmentations including color jittering. For the convenience of the experiment, we did not use the pose information for the V-COCO dataset. We perform all experiments with 3 NVIDIA RTX A6000 using Pytorch 1.9.0. We use batch size 11 per GPU for $\textrm{ViPLO}_{s}$ and 8 per GPU for $\textrm{ViPLO}_{l}$.

We use the AdamW \cite{loshchilov2017decoupled} optimizer for training with an initial learning rate of $10^{-4}$. For HICO-DET, we train the VIPLO for 8 epochs with flip data augmentation and the learning rate decay by a factor of 0.1. For V-COCO, we train the model for 20 epochs with additional data augmentations including color jittering, and decay the learning rate at the $10^{th}$ epochs. For the convenience of the experiment, we did not use the pose information for the V-COCO dataset. We perform all experiments with 3 NVIDIA RTX A6000 GPUs using the Pytorch 1.9.0. framework. We use batch size 11 per GPU for $\textrm{ViPLO}_{s}$ and 8 per GPU for $\textrm{ViPLO}_{l}$.

\section{Efficient computation for MOA}
\label{app:b}
The MOA module leads to a large performance increment in HOI detection, as shown in ablation studies in Sec. 4.3. But to use the MOA module, overlapped area $S$ has to be computed for each bounding box, which may be a computational burden under the CPU operation. So we design the entire process of computing $S$ to be possible through GPU operations. In specific, we compute the overlapped area of each row patch and column patch, then obtain the total overlapped area efficiently by multiplying these two. Details can be found in Algorithm \ref{alg:cap}.

\begin{algorithm}
\caption{Torch-like pseudo-code for the MOA module}\label{alg:cap}
\begin{algorithmic}[1]
\Input{box coordinate $\textbf{b}$, patch size $p$, attention map length $L$}
\Output{attention mask $A$ }
\State width = int($\sqrt{L}$)
\State $A$ = zeros($1, L$) 
\State $\textbf{b} = \textbf{b} / p$ 
\State $\textbf{b}_{int}$ = [floor($\textbf{b}$[0:2]), ceil($\textbf{b}$[2:4])]
\State $\textbf{b}_{wh}$ = 1 - abs($\textbf{b}_{int} - \textbf{b}$)
\State $a, b, c, d = \textbf{b}_{int}$
\State $x, y, z, w = \textbf{b}_{wh}$
\State row = arange (width * $b$ + $a$ + 1, width * $b$ + $c$ + 1)
\State mask\_index = row.repeat($d$ - $b$) + arange($d$ - $b$).repeat\_ interleave($c$ - $a$) * width
\State area\_row = [$x$, ones($c$ - $a$ - 2), $w$]
\State area\_column = [$y$, ones($d$ - $b$ - 2), $z$]
\State mask\_area = area\_row * area\_column
\State $A$[0, mask\_index] = mask\_area
\end{algorithmic}
\end{algorithm}

Another issue is that simply applying the MOA module increases the amount of computation in proportion to the number of regions in the given image. Hence, we propose three methods for reducing computation: 1) we apply the MOA module only in the last layer of ViT, which is sufficient for a feature to be conditioned to a given region; 2) we compute an attention score only for the CLS token, as the CLS token serves as an extracted feature; and 3) we calculate the dot product of the query and key only once, then add $\log(S)$ for each copied attention map in Eq. 1 in the paper. We apply three methods together to reduce the computational complexity of MOA. The computational complexity of the original ViT layer is $O(L^{2} \cdot C)$, and that of MOA applied the ViT layer is $O(M \cdot L \cdot C + M \cdot C^{2})$, where $L, C, M$ denotes the number of patches, hidden dimension, and the number of regions, respectively. The latter is linear to the number of patches since the number of regions is limited by non-maximum suppression (NMS), showing the efficiency of the MOA module.

\begin{figure}[t]
\begin{center}

\includegraphics[width=1.\linewidth]{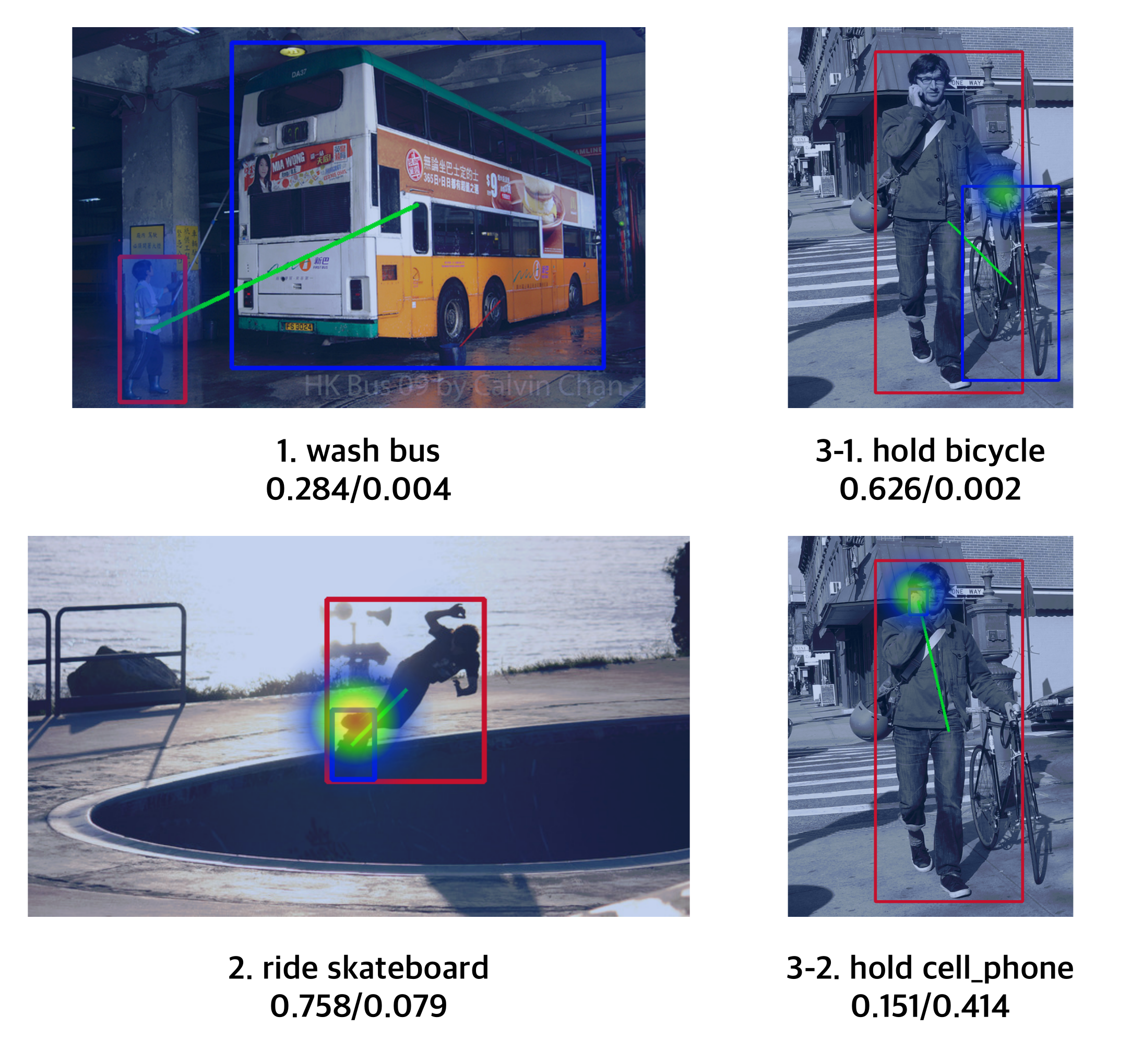}

\end{center}
\vspace{-5mm}
  \caption{Qualitative results of $\textrm{ViPLO}_{l}$ compared to the baseline SCG. For each image, the prediction scores of $\textrm{ViPLO}_{l}$ and SCG are shown (left: $\textrm{ViPLO}_{l}$, right: SCG). The joint attention in Eq.2 in the paper is also visualized as a heatmap.}
\vspace{-1mm}
\label{fig:qualitative}
\end{figure} 

\section{Qualitative Results}
\label{app:c}
We show qualitative results of ViPLO compared to the SCG in Fig. \ref{fig:qualitative}. We find that ViPLO can successfully detect difficult interactions where the human and object are far away (case 1). ViPLO also effectively detects interactions focusing on specific human joint, such as ankle or wrist in case of riding skateboard (case 2). Surprisingly, we find that our model focuses on different joints when detecting interaction for different objects, even in the same image (case 3). These results prove the effectiveness of ViPLO.

%\newpage
%\section*{Appendix}
%\include{Appendix}

\end{document}